\documentclass[journal]{article}
\usepackage{graphicx} 
\usepackage{booktabs} 
\usepackage{amsmath}

\usepackage{tikz}
\usetikzlibrary{shapes.geometric, arrows.meta, positioning}
\usepackage{amsmath}
\usepackage{amsfonts}
\usepackage{amssymb}
\usepackage{graphicx}

\title{Enes Causal Discovery}
\author{Alexis Kafantaris}
\date{March 2026}

\begin{document}

\maketitle
\section*{Abstract}

This paper is about a neural evolutionary method that is meant to address causal discovery. Furthermore, an interesting approach is implemented to train and evaluate a MoE and compare it to baseline causal discovery algorithms; this approach exploits the MoE's ability to learn both linear and nonlinear distributions of causal relationships among pairs of graph nodes and classify them. More specifically, the Enes(Edge node edge similarity) program, uses a mixture of experts neural net to determine the graph edge-node-edge triplet class from random non linear generated SEM; this way training is achieved in respect to causal pattern. The model inputs are graphs and their outputs are predictions regarding the triplet patterns and that is essentially the causal map for the problem, the penalties used are DAG enforcement penalty, and Pearson correlation penalty. In addition, this architecture is evaluated using the linear Pearson coefficient baseline as well as other state of the art solutions. Nonetheless, the primary focus is the sachs dataset, but then Michaelis-Menten dynamics is used in order to determine efficiency and scalability. Finally the model when compared with a statistical and a constrained programming baseline and it produces good results that indicate that it is performing in a stable and efficient way.
\newpage
\section{Introduction}

This paper is about a highly sophisticated architecture that addresses causal discovery using a neural net; the architecture is both robust and well performing. To begin with, causal discovery on sachs dataset is one interesting as well as classical problem. Causal discovery has many real life applications, such as finding the ground truth in service design processes. Nevertheless, starting it makes more sense to try and generalize from sachs data to other problems due to the complexity and the nature of it.

Determining the model is also important, as there are various tools that solve observational causal discovery and are also considered SOTA; most of them are not neural nets. In this particular instance, a neural net is used although it is rarely considered the best alternative due to the lack of formal training data. More specifically, causal discovery is observational and interventional; however, using random noise seems to suffice. Moreover, the model is a mixture of experts which has excellent modularizing capabilities.

In case of interventional data, the model would be trained to understand patterns within a given context of interventions. That is not the case, as here randomly generated data and some MM dynamics have been used in order to identify the sachs protein and other MM dynamics data. Furthermore, the model is used to classify the sachs data both real and artificial. The results are evaluated and, compared to some baselines, they are good. 

Lastly, the situation is analyzed and a brief discussion about the setup as well as the results explains the technical aspect of that venture. The paper is analyzed as follows, section two is the literature review, thereafter, the method and the model are described in section three, and the results are presented in section four, finally section five is the conclusion. 

\newpage
\section{Literature Review}
There are many programs used for causal discovery. Ranging from a simple Pearson coefficient\cite{margolin2006aracne}, to more complicated programs\cite{wang2023notears} to even more sophisticated neural nets\cite{lopez2017discovering}. An attempt is made every-time to determine the causal edges of a difficult and complicated graph\cite{sachs2005causal}. The graph describes a protein, a biological entity that is astutely organized by nature. Emerging are some forms of relationships, i.e. causal relationships, and these can be further examined; moreover, causal relationships can be classified to determine which edges are related and which are not.

To achieve causal discovery, one needs to provide a graph as input and determine some relationships. And the observational approach has been used, as the main focus is the Karen Sachs benchmark\cite{sachs2005causal}. There are other objectives in mind, like discovering the silver standard ground in service design datasets but these are secondary. In other words, this model does not know anything about the real data, so it uses or some math in order to determine causal relationships among the nodes-edges. 

Additionally, many other approaches have been implemented, especially statistical algorithms like PC\cite{spirtes1991algorithm}, or mathematical algorithms like NOTEARS\cite{wang2023notears}. From these and others, it seems like the best alternative is a fast tool with no induction base. Although, there are notably machine learning models and neural networks for causal discovery, presumably both interventional and observational\cite{lopez2017discovering}, there are few specific systems\cite{lorch2021amortized}\cite{dagpa2025}. Moreover, only recently\cite{dagpa2025} has the induction base been addressed using the notears formula and a neural net.

One of the major issues that arise is the data; usually the saying garbage in garbage out refers to a model where the training occurs with poor data. However, in the scheme of observational causal discovery the GIGO is unavoidable. For this reason a point is made to obtain at least partially useful induction basis, due to the fact that an induction basis has its merits\cite{dagpa2025}. Here training a neuron for data similarity makes absolute sense as according to the actual experiment\cite{sachs2005causal} there are similar motifs to be found.

Furthermore scaling is an issue\cite{wang2023notears}; although the notears method for works it struggles due to the mathematical overhead. While on the other hand methods like PC\cite{spirtes1991algorithm} an induction basis and hence scale but can be easily confused due to their static nature. However, recently bridging this gap is the dagpa system\cite{dagpa2025} that uses the trace objective and can be both scaled and perform well in real datasets. A thorough system, however, would be better; when it comes to systems that are explainable, where the MoE architecture modularizes everything, and is completely transparent. 

The proposed architecture is a mixture of experts, which allows the model entities, such as causal relationships, to be further parameterized\cite{shazeer2017outrageously}. More specifically, an attempt is made to exploit a neural net, as implementing neurons poses a great challenge for this dataset. To explain, a simple and fast Pearson coefficient linear model usually achieves good scores. An aggressive baseline that requires a really good model to overcome that is. Moreover, there are major limitations when it comes to causal discovery of observational data. Unlike the sachs\cite{sachs2005causal} one did not use interventions but only prior knowledge; the most prohibitive limitation is that of the data which is addressed.

\section{Methodology}
As mentioned above the proposed architecture is a mixture of experts physics informed; a physics informed network\cite{raissi2019pinn} is a network that uses physical constraint as it's core learning mechanism. Moreover, to achieve the objective, these constraints are merged in the training matrix. In addition, this network has also penalized the constraint as it was found to produce more stable and robust results this way\cite{shazeer2017outrageously}. There are three primary similarities, which govern the physics of the network, namely cosine similarity, Pearson coefficient, and adjacency matrix constraint. The model is, therefore, trying actively to minimize the similarities, in both a linear and a non linear fashion, of an edge-node-edge probe of the data. 

Additionally, the network uses fusion as its core matrix operation of these entities with the loss function; there is also a gating mechanism for the experts, which suggests that each learned class has both linear and non-linear traces. A lot of thought has been put into this model, especially to avoid triviality, i.e. a good working case with underlying causal collapses on others. In other words, the model is meant to understand patterns at a low level subgraph and compare the similarities like a polynomial equation would. Except for the fusion and the gating mechanism, self supervised penalty weighting is put into place. Theoretically, it should not differ from static weighting and in practice it seems to achieve earlier stabilization of model parameters. For the experiments purposes a unit weight was used due to the fact that the data were sub optimal; to use self supervised weights one needs to have an intimate understanding of the data to, and then one can fine tune the model to underestand his data easier. 

One of the key advantages of this method is the induction basis; induction allows faster inferences and scalability. However, there is a saying : garbage in garbage out. Regarding causal discovery, one is confronted by GIGO when it comes to training Enes. On the one hand, using relevant data is like cheating because the model knows where the patterns come from, which is interventional\cite{sachs2005causal}. On the other hand, using other data is confounded; at least that is how observational classification has to be. Moreover, shapes and forms are generated and a middle ground of some MM kinetics is also employed\cite{michaelis1913}. These are not fully adopted to maintain the unbiased perspective of the program.

Lastly, a case of stability is made during the training; the program is already confronted by GIGO. There are also multiple objectives at hand and a complicated fusion loss. In order to operate at its peak capacity stabilization during its training would really help. For this reason Langevin dynamic and simulated annealing optimization\cite{vargas2025langevin} are employed. This way the chaotic gradient descent slope becomes more stable and more predictable allowing for higher precision. A classifying task is at hand. The model is set up and the hyper-parameters are optimized; it is a fast and a stable model called Enes. Now predictions will be examined for both the sachs dataset and the MM kinetics, without changing the unbiased data and using fully MM generator.

Following is a drawing of the neural net architecture:
\newpage
 \begin{figure}[ht]
\centering
\begin{tikzpicture}[
    node distance=1.0cm,
    io/.style={draw, rectangle, rounded corners, fill=blue!10, minimum width=8.5cm, minimum height=0.7cm, align=center},
    process/.style={draw, rectangle, fill=orange!10, minimum width=8.5cm, minimum height=0.7cm, align=center},
    expert/.style={draw, rectangle, fill=green!5, minimum width=3.8cm, minimum height=1.0cm, align=center, font=\scriptsize},
    penalty/.style={draw, rectangle, dashed, fill=red!5, minimum width=3.2cm, minimum height=0.7cm, align=center, font=\tiny},
    arrow/.style={->, thick, >=stealth}
]

\node (inputs) [io] {Input: Raw Observational Triplet $X_{i,j,k} \in \mathbb{R}^{3 \times T}$};
\node (cnn) [process, below=1.2cm of inputs] {Shared Feature Extraction };

\node (expert1) [expert, below left=0.8cm and -0.5cm of cnn] {Expert A: \\ Non linear Expert };
\node (expert2) [expert, below right=0.8cm and -0.5cm of cnn] {Expert B: \\ Linear Expert};

\node (gate) [process, below=3.8cm of cnn] {Expert Gating Mechanism \\ $Score = w \cdot A + (1-w) \cdot B$};
\node (class) [process, below=of gate] {Softmax Classification };
\node (out) [io, below=of class] {Output: Causal Motif Prediction};

\node (topo) [penalty, left=0.8cm of gate, yshift=1.0cm] {Adjacency Constrain \\ $b*W^2$ Penalty};
\node (faithful) [penalty, left=0.8cm of gate, yshift=0.1cm] {Pearson correlation \\ $a*P(edge) \cdot (1-|r|)$};
\node (cosim) [penalty, left=0.8cm of gate, yshift=-0.8cm, fill=purple!5] {Cosine Similarity \\ $c*\cos(\theta_{A,B})$ Consistency};

\draw[arrow] (inputs) -- (cnn);

\draw[arrow] (cnn.south) -- ++(0,-0.3) -| (expert1.north);
\draw[arrow] (cnn.south) -- ++(0,-0.3) -| (expert2.north);

\draw[arrow] (expert1.south) |- ++(0,-0.8) -| ([xshift=-2.5cm]gate.north);
\draw[arrow] (expert2.south) |- ++(0,-0.8) -| ([xshift=2.5cm]gate.north);

\draw[arrow, dashed, red!70!black] (topo.east) -- ([yshift=0.5cm]gate.west);
\draw[arrow, dashed, blue!70!black] (faithful.east) -- ([yshift=0.1cm]gate.west);
\draw[arrow, dashed, purple!70!black] (cosim.east) -- ([yshift=-0.4cm]gate.west);

\draw[arrow] (gate) -- (class);
\draw[arrow] (class) -- (out);

\end{tikzpicture}
\caption{Final architecture of the Enes Model. The model is comprised of a dual expert architecture and graph metrics that are treated as the physics of the network. It inputs graph nodes and outputs causal pattern classification.}
\end{figure}
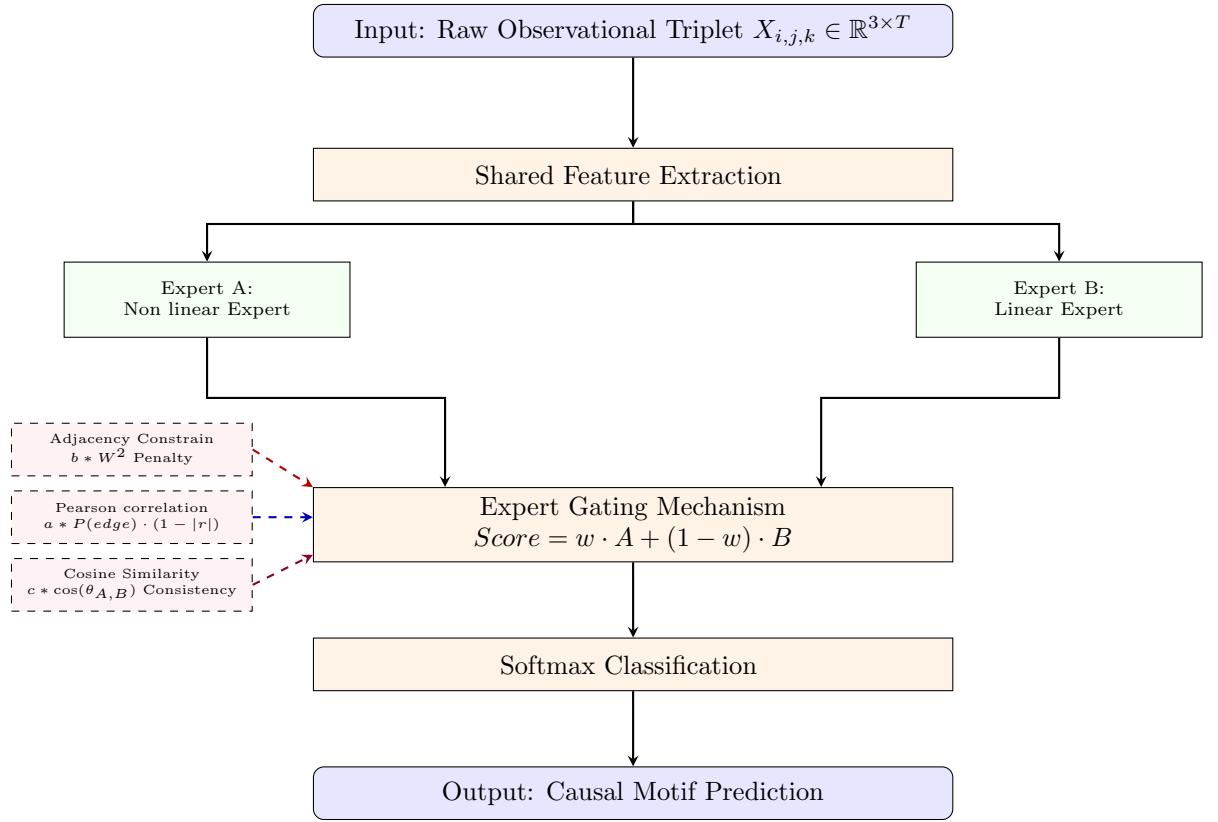

\section{Results and Discussion}
The Enes model classifies edge node edge similarity based on GIGO data and some MM that it has received during the training. The gigo can t get better without cheating. For the evaluation part some the following metrics are defined; 

\noindent \textbf{Evaluation Metrics}
\begin{align*}
    \text{TP} &= \sum_{i,j} I(\hat{W}_{i,j} = W_{i,j} \neq 0) \\
    \text{Prec} &= \frac{\text{TP}}{\sum I(\hat{W}_{i,j} \neq 0)}, \quad \text{Rec} = \frac{\text{TP}}{\sum I(W_{i,j} \neq 0)}, \quad \text{F1} = \frac{2 \cdot \text{Prec} \cdot \text{Rec}}{\text{Prec} + \text{Rec}} \\
    \text{SHD} &= \sum_{i < j} I(\hat{W}_{i,j} \neq W_{i,j} \lor \hat{W}_{j,i} \neq W_{j,i}) \\
    \text{Acc} &= \frac{d^2 - \text{SHD}}{d^2}, where \ d \ is \ the\ matrix \ dimension.
\end{align*}
Each metric has its own significance; for causal discovery, one cares a lot about SHD and precision. The lower the SHD the less mistakes it makes per graph, while as for precision, the higher the precision is the better a model is at picking up correct edges. Therefore, one cares about accuracy to; it can be stated that the TP and the recall are not so important after all as the model does not know the data a priori.

Following is the benchmark evaluation of the results:
\begin{table}[ht]
\centering
\noindent \textbf{Table 1: Performance and Scalability Metric Benchmark}
\resizebox{1.0\textwidth}{!}{
\begin{tabular}{lcccc}
\hline
\textbf{Metric} & \textbf{11 Nodes (Sachs)} & \textbf{11 Nodes (MM)} & \textbf{25 Nodes (MM)} & \textbf{50 Nodes (MM)} \\
\hline

\multicolumn{5}{c}{\textbf{Linear Fast Pearson}} \\
Accuracy  & 0.8694 & 0.8107 & 0.9526 & 0.9789 \\
Precision & 0.6223 & 0.5091 & 0.1889 & 0.2966 \\
Recall    & 0.3294 & 0.2605 & 0.0708 & 0.0490 \\
F1 Score  & 0.4308 & 0.3444 & 0.1030 & 0.0827 \\
SHD       & 15.80  & 22.91  & 29.60  & 52.80  \\

\hline
\multicolumn{5}{c}{\textbf{PC Algorithm }} \\
Accuracy  & 0.7025 & 0.8512 & 0.9360 & 0.9704 \\
Precision & 0.0870 & 0.3333 & 0.3621 & 0.3711 \\
Recall    & 0.1176 & \textbf{0.8033} & \textbf{0.8710} & \textbf{0.7350} \\
F1 Score  & 0.1000 & \textbf{0.4706} & \textbf{0.5122} & \textbf{0.4932} \\
SHD       & 36.00  & 18.00  & 40.00  & 74.00  \\

\hline
\multicolumn{5}{c}{\textbf{Enes}} \\
\hline
\textbf{Accuracy}  & \textbf{0.8818} & \textbf{0.8240} & \textbf{0.9614} & \textbf{0.9802} \\
\textbf{Precision} & \textbf{0.7214} & \textbf{0.6565} & \textbf{0.5333} & \textbf{0.5000} \\
\textbf{Recall}    & \textbf{0.2706} & 0.2193          & 0.2834          & 0.1347          \\
\textbf{F1 Score}  & 0.3939          & 0.3296          & 0.3598          & 0.2503          \\
\textbf{SHD}       & \textbf{14.30}  & \textbf{21.29}  & \textbf{23.00}  & \textbf{49.60}  \\
\hline
\end{tabular}
}
\caption{Performance comparison. Precision, accuracy, recall, F1, and SHD are evaluated. Three programs are used: Enes, PC, and linear pearson coefficient as an objective baseline. Lastly, the evaluation is performed upon the real sachs data and synthetic MM kinetics that are generated for three sizes.}
\end{table}
After a thorough evaluation, several good things can be seen. The model clearly outperforms the linear baseline and the constrained based baseline in most respects. Clearly, the model has better SHD which makes it more robust for observational causal discovery. Make no mistake, both of these programs are highly performing baselines especially on their own accord, and achieving a better result is certainty difficult. Furthermore,  the recall index is not higher, which suggests that the Enes Model cherry picks results.

The most interesting observation is the way the systems plateau; here it is noted that for observational data classification these baselines are also considered state of the art. The metrics provided are well beyond expectations, especially for the real dataset. Considering the gigo premise that is, as one would expect garbage outputs from the model. This was not the case, it is attested that the data generation was attempted using only patterns for which again the model performed similarly. However, data generation closer to other systems was preferred due to its faster training potential. It is argued that random linear and non linear structural equation model data, random patterns, and noise is not eventually bad data.

Another interesting note is about the baseline; historically the observational sachs data seem to be modeled better using simple and linear methods\cite{margolin2006aracne}.  So, one would indeed expect neural networks to perform well, but they are not. Many architectures were tested, including convolutional neural nets, all of which exhibited the same trend; networks usually had higher accuracy but lower recall and F1 score. However, the difference is more than noticeable, as this architecture was proved the best by empirical validation. Despite the lack of high results in the MM dynamics, it still performed well enough. 

\section{Conclusion}

It is noted here that this is a hybrid system, one can just change the training data, and the model becomes fully tuned to the actual problem. That is another topic, however, and it was avoiding because if that was the case the model would be cheating. By using random generated data and preserving it through the experiments the actual capabilities of the model were addressed for its capacity to generalize based on similarity. One could also do a polynomial or a mutual information comparison, and that did not come close to the baseline despite being sound. Essentially, one replaced this polynomial with the global estimator neural net and that worked.

Finally, it is important to address the elephant in the room, which is the sachs protein dataset, byt the real reason this model exists is service design processes ground truth retrieval. Having the ability to classify proteins is the hardest causal discovery task, and if the model performed well enough in this task, it would perform well enough in other tasks. Especially when the data was tweaked; for this reason the MM dynamics evaluations were tested, but yet again this could much better.
\newpage
\section*{Acknowledgments}
It is acknowledged that this paper was written to address the research venture of my PhD thesis in Aueb. The PhD thesis is about fuzzy optimization of information transmission of service design processes.

\end{document}